\pdfoutput=1

\documentclass[11pt]{article}

\usepackage[final]{acl}

\usepackage{times}
\usepackage{latexsym}

\usepackage{booktabs}
\usepackage{listings}
\usepackage{color}
\usepackage{caption} 

\definecolor{gray}{rgb}{0.5,0.5,0.5}
\definecolor{blue}{rgb}{0.13,0.13,1}

\usepackage[T1]{fontenc}

\usepackage[utf8]{inputenc}

\usepackage{microtype}

\usepackage{inconsolata}

\usepackage{graphicx}

\title{Reflect before Act: Proactive Error Correction in Language Models}

\author{Qiuhai Zeng \\
  Pennsylvania State University \\
  \texttt{qiuhai.stat@gmail.com} \\\And
  Sarvesh Rajkumar \\
  Amazon \\
  \texttt{rajksarv@amazon.com} \\\And
  Di Wang \\
  Amazon \\
  \texttt{imdi@amazon.com} \\\AND
  Narendra Gyanchandani \\
  Amazon \\
  \texttt{gyanchan@audible.com} \\\And
  Wenbo Yan \\
  Amazon \\
  \texttt{yanwenb@amazon.com} }

\begin{document}
\maketitle
\begin{abstract}
Large Language Models (LLMs) have demonstrated remarkable capabilities in interactive decision-making tasks, but existing methods often struggle with error accumulation and lack robust self-correction mechanisms. We introduce "Reflect before Act" (REBACT), a novel approach that enhances LLM-based decision-making by introducing a critical reflect step prior to taking the next action. This approach allows for immediate error correction, ensuring smooth action path and adaptibity to environment feedback. We evaluate REBACT on three diverse interactive environments: ALFWorld, WebShop, and TextCraft. Our results demonstrate that REBACT significantly outperforms strong baselines, improving success rates by up to 24\% on WebShop (achieving 61\%), 6.72\% on ALFWorld (achieving 98.51\%), and 0.5\% on TextCraft (achieving 99.5\%) using Claude3.5-sonnet as the underlying LLM. Further analysis reveals that REBACT's performance improvements are achieved with only a few modification steps, demonstrating its computational efficiency.
\end{abstract}

\section{Introduction}
Large Language Models (LLMs) have demonstrated proficiency in a variety of complex tasks \cite{wei2022chain, huang2022towards}. They have been increasingly applied to interactive decision tasks \cite{yao2022react, qin2023toolllm}. Methods have focused on how to determine the optimal next action \cite{yao2022react, yao2024tree, hao2023reasoning, zhou2023language}, handling mistakes in forthcoming actions \cite{prasad-etal-2024-adapt, schroeder2024thread}, and reflecting the complete sequence of actions \cite{shinn2024reflexion, madaan2024self}. However, less attention has been given to correcting errors in previously executed actions during the course of action.

In this work, we present \textbf{Re}flec \textbf{b}efore \textbf{Act} (REBACT), an iterative self-reflection approach designed for LLMs engaged in interactive decision-making. This approach is designed to identify and rectify errors in previously executed actions before proceeding to the next. Within a single LLM invocation, REBACT generates modified previous actions (if necessary) along with the next action. to detect and rectify errors in previous executed actions before executing the next action. Within one LLM call, REBACT generated both modified previous actions if any and the next action. If a modification is needed, REBACT will execute the corrected action; otherwise, it continues with the next planned action as usual.

We conducted experiments using three different datasets: (1) WebShop \cite{yao2022webshop}, where LLMs need to purchase products on a shopping website, (2) ALFWord \cite{shridhar2020alfworld}, where LLMs need to interact with a virtual household to complete tasks, and (3) TextCraft \cite{prasad-etal-2024-adapt}, where LLMs craft recipes using specific commands. Across all three datasets, REBACT outperformed strong baselines, improving success rates by 24\% on WebShop (achieving 61\%), 6.72\% on ALFWorld (achieving 98.51\%), and 0.5\% on TextCraft (achieving 99.5\%).

\begin{figure}[t]
  \includegraphics[width=\columnwidth]{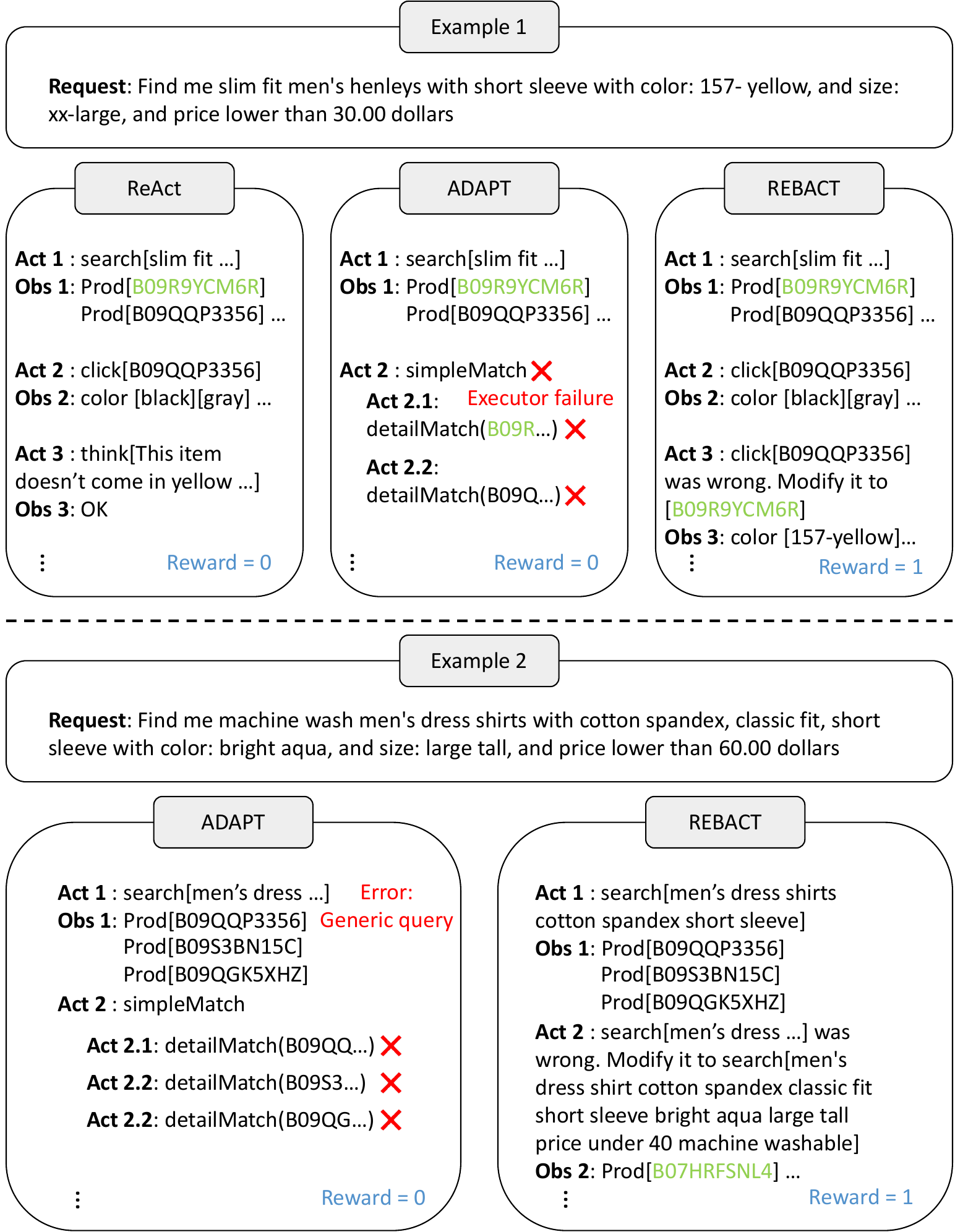}
  \caption{Comparison of REBACT and baselines ReAct and ADAPT on two WebShop examples. In both examples, we only show task solving trajectories generated by the model (Act) and the environment (Obs).}
  \label{fig:Example}
\end{figure}

\section{Related work}
\textbf{Decision-making} Prior works such as SayCan \cite{brohan2023can} and WebGPT \cite{nakano2021webgpt} are 'act-only' methods that sequentially generate the next action during environmental interaction, relying solely on the intrinsic capability of the Language Model (LLM) to adapt. In contrast, ReAct \cite{yao2022react} introduces an "act-reason" approach by generating reasoning traces alongside action generation, aiding the LLM in adapting to environmental feedback. However, ReAct lacks mechanisms to handle errors in actions, as shown in Figure~\ref{fig:Example} Example 1. ADAPT \cite{prasad-etal-2024-adapt} employs a recursive decomposition strategy to address failed tasks by decomposing them into sub-tasks with a planner expert's assistance. However, it depends on the executor expert to generate success heuristics for atomic skills, as illustrated in Example 1 of Figure 1. Additionally, ADAPT lacks a clear mechanism to correct errors in previously successful actions, which can accumulate and lead to task failure as more feedback is received from the environment. Reflexion \cite{shinn2024reflexion} introduces the ability to detect action errors but only after task completion, and needs to convert binary or scalar feedback into verbal feedback for LLM agents to adjust their decision-making strategies. In this paper, we show that by engaging in self-reflection at every decision point before proceeding to the next action, the LLM agent can effectively identify and correct previous errors, ensuring a smooth and successful action path. 

\section{REBACT}
\begin{figure}[t]
  \includegraphics[width=\columnwidth]{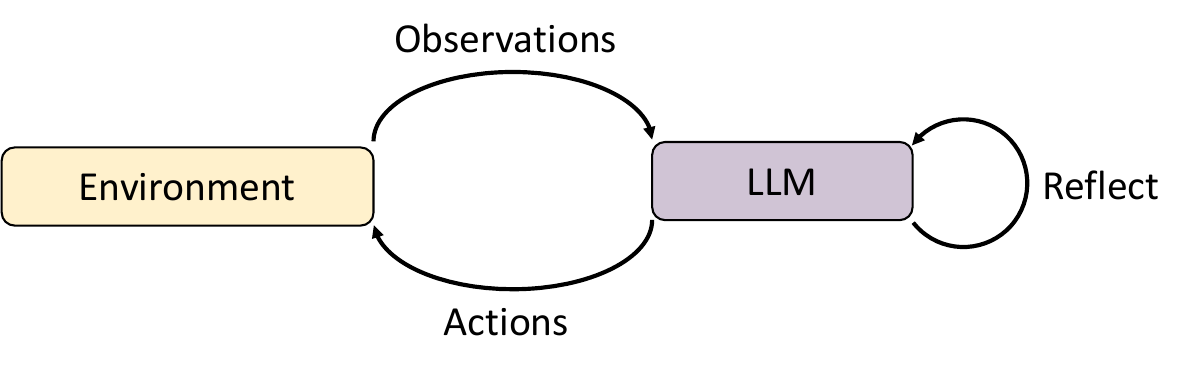}
  \caption{Diagram of the REBACT method.}
  \label{fig:Diagram}
\end{figure}

We present REBACT, a novel approach to enhance interactive decision-making through continuous reflection. At each decision point, REBACT prompts LLMs to evaluate the need for modifying any previously executed actions (Figure~\ref{fig:Diagram}). The method incorporates two common elements in its prompts: successfully completed tasks and current action-observation pairs related to the ongoing human request. The LLM is then instructed to determine if any previous actions require adjustment, produce necessary changes, if any, and formulate the next action. The decision process follows a simple rule: If an action is adjusted, execute the revised version; Otherwise, proceed with the planned action. As shown in Figure~\ref{fig:Example}, REBACT successfully detects errors in previously executed actions based on feedback from the environment and modifies the actions to correct them. This approach offers two advantages: (1) Continuous reflection mechanism ensures the action strategy including previous actions has adapted to the environment. (2) By conducting both reflection and planning for next steps within the same LLM call, REBACT achieves efficient integration of the decision-making process.

\section{Experiments and Results}
We used Claude3.5-sonnet to conduct all our experiments. In the following we describe the datasets used in our experiments and baselines used for comparison with REBACT. 

\subsection{Datasets}
In this paper, we investigate the effectiveness of REBACT across three diverse datasets.

First, WebShop simulates an online shopping experience with 1.18 million real-world products and requires agents to navigate the website and purchase products to meet specific user requests. Following \cite{shinn2024reflexion, prasad-etal-2024-adapt}, we assess our models on 100 user instructions. Second, ALFWorld \cite{shridhar2020alfworld}, a text-based adaptation of the ALFRED benchmark \cite{shridhar2020alfred} situated within the TextWorld environment, features 6 distinct task types where agents navigate and interact through text-based commands to accomplish tasks such as placing a clean mug on a desk. Following \cite{shridhar2020alfworld, prasad-etal-2024-adapt}, our study presents the results on 134 unseen evaluation games. Finally, TextCraft \cite{prasad-etal-2024-adapt} introduces a text-only crafting environment inspired by Minecraft\footnote{\url{https://www.minecraft.net}}. Here, agents engage in tasks of cooking recipes with a natural compositional structure, like crafting beehive using oak planks and honeycomb. Following \cite{prasad-etal-2024-adapt}, we evaluate the approach on a test set of 200 tasks.

\subsection{Baselines}
Our study compares REBACT with two established approaches. ReAct, introduced by \cite{yao2022react}, alternates between generating reasoning traces and actions to interact with the environment. ADAPT, developed by \cite{prasad-etal-2024-adapt}, breaks down failed tasks into manageable sub-tasks for successful execution. For both approaches, we used prompts from their original studies. To account for differences in our underlying model, we added specific instructions to guide the LLM in generating output in the expected format.

\subsection{Results}
We show that REBACT achieves the highest success rate compared to baselines from prior work on ALFWorld, WebShop, and TextCraft datasets.

\textbf{WebShop} In Table~\ref{tab:table1}, we show that REBACT achieves the highest success rate across the board, while using ReAct alone results in the lowest overall performance. Through reflecting on the previous actions, REBACT improves success rate by 24\% compared to ADAPT and 32\% compared to ReAct. As reported by \cite{yao2022webshop}, the human (expert) got a success rate of 59.6\% and the human (average) got a success rate of 50 on all 500 user instructions. Despite being tested on a smaller subset of 100 user instructions, REBACT achieved a higher success rate compared to both expert and average human participants. This demonstrates REBACT's potential to outperform human experts in selecting the correct products based on user requests within this dataset. We also calculated the average score for comparison and found that REBACT yielded the highest score compared to the other two baselines, outperforming ADAPT by 35.71 and ReAct by 42.56. Additionally, REBACT also had higher score than the average human.

\begin{table}
\centering
\begin{tabular}{lrr} 
\toprule
\textbf{Method} & \textbf{Score} & \textbf{Success Rate} \\ 
\midrule
ReAct            & 33.62                     & 29                               \\
ADAPT            & 40.47                     & 37                               \\
REBACT           & \underline{76.18}                     & \textbf{61}                               \\
Human (expert)*  & \textbf{82.1}                      & \underline{59.6}                             \\
Human (average)* & 75.5                      & 50                               \\
\bottomrule
\end{tabular}
\caption{REBACT achieves the highest success rates (\%) on WebShop. Best (highest) success rates are highlighted in bold and second-highest rates are underlined. *Success rate reported by \cite{yao2022webshop}. } \label{tab:table1}
\end{table}

\textbf{ALFWorld} REBACT achieves the highest overall success rate as well as the highest success rate across all task types (Table~\ref{tab:table2}). Specifically, REBACT improved upon ReAct's overall performance by 6.72\% and ADAPT's by 14.93\%. Among 6 distinct task types, REBACT got a 100\% success rate on four types of tasks. Notably, in the Clean and Pick2 (pick two items) tasks, where both ReAct and ADAPT got success rates below 85\%, REBACT achieved a 100\% success rate. 

\textbf{TextCraft} Table~\ref{tab:table3} shows that REBACT achieves the highest rate at 99.5\%. REBACT's success rate is 0.5\% higher than ADAPT's, a margin considered significant as both scores are close to 100. This highlights that simply correcting errors in previous actions is sufficient to successfully complete tasks with compositional structures, and even outperforms the idea of recursive decomposition of tasks based on those structures. Notably, REBACT's performance significantly exceeds that of ReAct by 19.5\%, given that ReAct lacks an error correction feature and only achieves an 80\% success rate.

\begin{table*}[h]
\centering
\begin{minipage}{0.6\textwidth}
\centering
\begin{tabular}{l|cccccc|c}
\toprule
\textbf{Method} & \textbf{Pick} & \textbf{Clean} & \textbf{Heat} & \textbf{Cool} & \textbf{Examine} & \textbf{Pick2} & \textbf{All} \\
\midrule
ReAct & \textbf{100} & 83.87 & 91.3 & \textbf{95.24} & \textbf{100} & 82.35 & 91.79 \\
ADAPT & 95.83 & 83.87 & 91.3 & 85.71 & 55.56 & 82.35 & 83.58 \\
REBACT & \textbf{100} & \textbf{100} & \textbf{95.65} & \textbf{95.24} & \textbf{100} & \textbf{100} & \textbf{98.51} \\
\bottomrule
\end{tabular}
\caption{REBACT consistently achieves the highest success rates (\%) across all types of tasks, as well as in overall performance, on ALFWorld.}
\label{tab:table2}
\end{minipage}%
\hfill
\begin{minipage}{0.25\textwidth}
\centering
\begin{tabular}{lr}
\toprule
\textbf{Method} & \textbf{SR} \\ 
\midrule
ReAct  & 80 \\
ADAPT  & 99 \\
REBACT & \textbf{99.5} \\
\bottomrule
\end{tabular}
\caption{REBACT achieves the highest success rates (\%) on TextCraft.}
\label{tab:table3}
\end{minipage}
\end{table*}

\section{Analysis}
We analyze REBACT approach in terms of its efficiency and adaptability.

\subsection{Does REBACT require many LLM calls?} \label{sec:LLMCalls}
\begin{figure}[t]
  \includegraphics[width=\columnwidth]{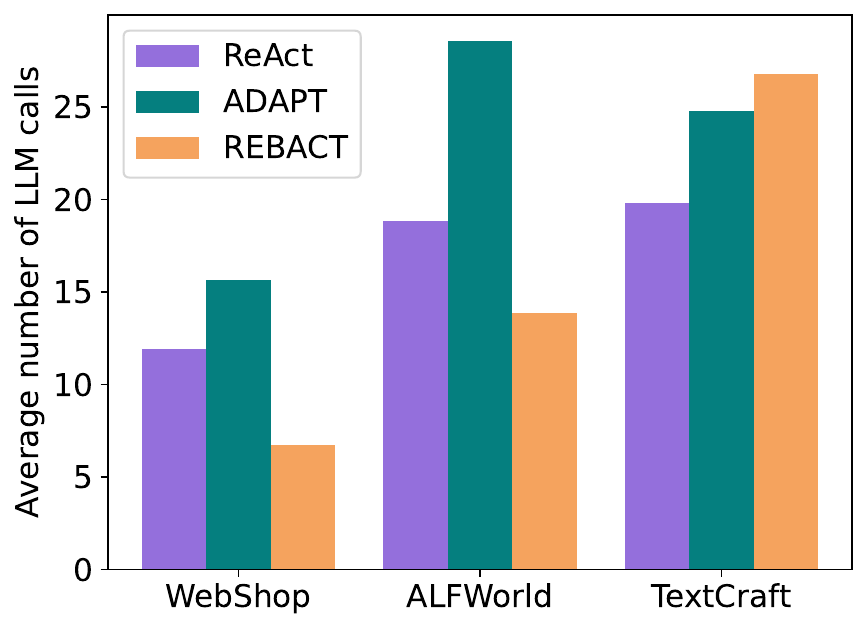}
  \caption{Average number of LLM calls across tasks for REBACT and baselines across datasets.}
  \label{fig:LLMCall}
\end{figure}

The effectiveness of decision-making agents is shown to increase with the number of permissible calls to an LLM \cite{prasad-etal-2024-adapt}. To determine if the enhancements seen with REBACT stem from this factor, we calculated the number of LLM calls across the WebShop, ALFWorld, and TextCraft datasets. Figure~\ref{fig:LLMCall} shows that on WebShop and ALFWorld, REBACT requires the fewest number of LLM calls, achieving reductions of 57\% and 26\%, respectively, compared to the next best performing methods (ADAPT for WebShop and ReAct for ALFWorld). On the TextCraft dataset, the number of REBACT calls is very similar to the number of ADAPT calls. The findings demonstrate the computational efficiency of the REBACT method, primarily because the reflection process in REBACT is integrated with the generation of subsequent actions. 

\subsection{How many modifications are needed?}
\begin{figure}[t]
  \includegraphics[width=\columnwidth]{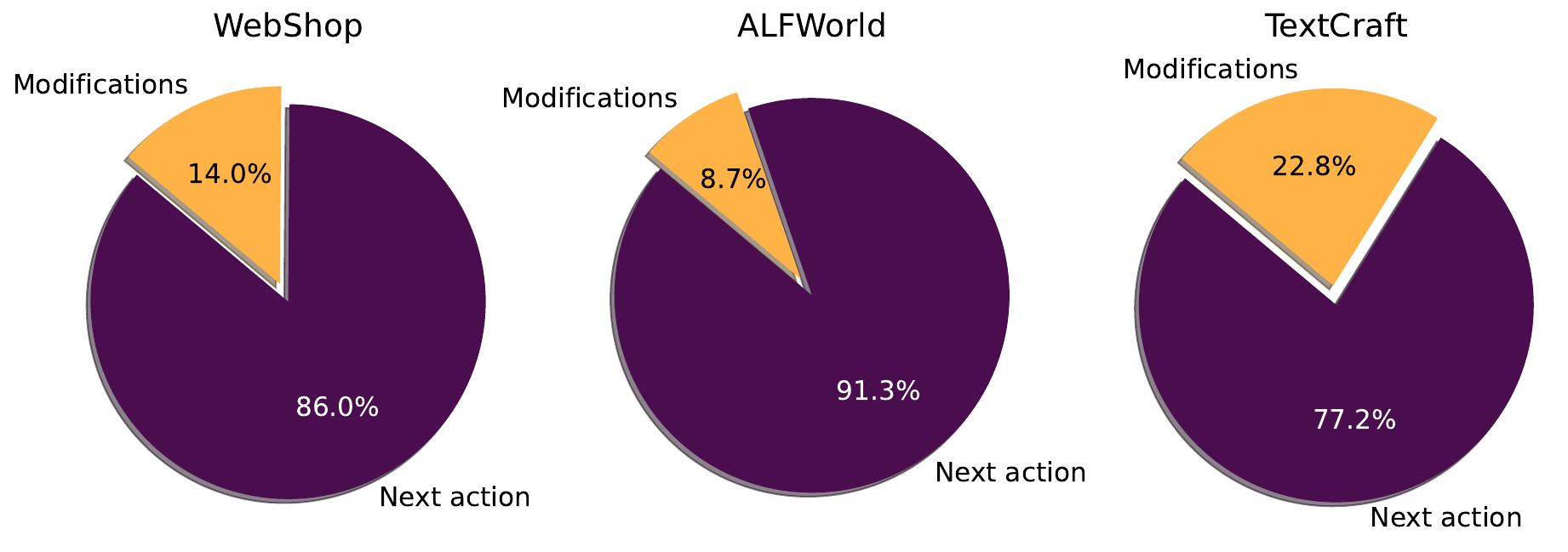}
  \caption{Proportion of LLM calls used for modifying incorrect actions and executing next actions for REBACT across datasets.}
  \label{fig:ModifyNum}
\end{figure}

A critical component of the REBACT approach involves allowing the LLM to execute modified actions to rectify errors in previous actions, significantly enhancing its performance. We analyze what proportion of LLM calls are dedicated to these modifications. As shown in Figure~\ref{fig:ModifyNum}, this proportion ranges from 8.7\% in ALFWorld to 22.8\% in TextCraft. The relatively low percentage of calls used for modifications demonstrates the LLM's capability to effectively adapt to subtle environmental changes and facilitate continuous adjustments and enhancements throughout the action process. The necessity for a higher proportion in TextCraft arises because the LLM often attempts to directly obtain the target item and/or ingredients from the environment when it is required to craft them using other items. REBACT effectively addresses these failed attempts, though it requires a few LLM calls to do so, resulting in a total number of LLM calls comparable to that of ADAPT, as detailed in Section~\ref{sec:LLMCalls}.

\section{Conclusion}
We propose REBACT, an iterative self-reflection strategy designed for interactive decision-making in LLMs. Unlike existing methods that focus on generating and executing the subsequent action, REBACT emphasizes reflecting and possibly revising past actions to address any allowable errors before proceeding. Our simple method not only outperforms others in success rates across the WebShop, ALFWorld, and TextCraft datasets but also demonstrates computational efficiency. Moreover, REBACT highlights the potential of LLMs to self-reflect on their own decisions given environment feedback, encouraging further exploration of the ability to introspect in non-decision-making tasks such as conversation.

\section{Limitations}
REBACT relies on the LLM to detect and correct the error made in the previous actions. However, not all actions can be modified or corrected in real-world applications. For example, once emails are sent, we can no longer edit the content and resend the emails. A viable solution could be to send a follow-up email addressing any inaccuracies in the previous emails. Therefore, we need to be cautious when identifying modifiable actions and formulating appropriate corrective strategies. Additionally, REBACT's effectiveness is contingent on receiving environmental feedback to facilitate its reflective processes \cite{huang2023large}.

\bibliography{main}

\appendix

\section{Prompts}
\label{sec:appendix}

\centering{\textbf{Prompt: REBACT on WebShop}}
\begin{lstlisting}
You are currently on the search results page. You will be given the current request of purchasing a product, the previous search action, and the most recent observation from the website.
Your task is to 
1. decide whether any of the previous {} actions (see below) needs to be modified. 
2. decide the most appropriate next action. 

When you reply, you must follow these rules:
1. If you think any previous actions is wrong and want to modify the previous action, you must provide the modified action that is different from the previous action. If the modified action is a click action, you can only click on options enclosed in square brackets [] from the closest observation before that previous action.
2. When you determine the next action, you can only click on text enclosed in '[]' from the most recent observation after current request.
3. The action could only be either 'click' or 'search'. You must replace the 'action' word in the following reply template with either 'click' or 'search'.

You must reply in this format:
"Previous action {} is [correct or wrong]. This action should be modified to: action[...]
The next action is: action[...]"

<a successful example>
<current observation-action pairs>
\end{lstlisting}

\centering{\textbf{Prompt: REBACT on TextCraft}}
\begin{lstlisting}
You can perform the following actions to interact with the environment: 
- craft [target count] [target item] using [count] [item]
- get [count] [item]
- inventory
Here [count] is a place holder for number of object, and [item] is placeholder for name of object.

Please decide whether the previous action '{}' needs to be modified and decide the most appropriate next action. If the previous action needs to be modified, make sure the modified action is different from the previous action and it is executable. The next action should be different from the modified action. If the previous action does not need to be modified, repeat the previous action as the modified action.
Please reply in this format: "Previous action '{}' is [correct or wrong]. It should be modified to: [action].
The next action is: [action]." 

You are given few useful crafting recipes to craft items in Minecraft. Crafting commands are of the format "craft [target object] using [input ingredients]". You can either "fetch" an object (ingredients) from the inventory or the environment or "craft" (target) using any of the crafting commands. You can use ONLY these crafting commands provided, do not use your own crafting commands. However, if the crafting command uses a generic ingredient like "planks", you can use special types of the same ingredient e.g. "dark oak planks" in the command instead. For any other natural language or thoughts, use prefix 'think: '.

Here is a demo of how to fetch and craft objects.

<atomic examples>
Here is an example of a complex goal.
<React example>
Here is a different goal with different craft commands. You can take the help of crafting commands below to create new objects. Keep in mind that:
- It is okay to generate more target objects than your goal.
- Be very careful with the count of objects, SAME object counts mentioned in the input crafting command. 
- You cannot use a partial crafting command recipe, i.e. if the recipe generates 2 objects you CANNOT alter it to produce just 1. 
- Also, you can use ONLY 1 crafting command in your plan.
<new task description>
<current observation-action pairs>
\end{lstlisting}

\centering{\textbf{Prompt: REBACT on ALFWorld}}
\begin{lstlisting}
You are a helpful robot navigating through a household. Your task is related to some of the following tasks:
- Put an item in/on a receptacle
- Take an item from a receptacle
- Heat an item with a receptacle
- Cool an item with a receptacle
- Clean an item with a receptacle
- Use a desklamp to look at an item
You can only hold one item in your hand. If you have previously taken an item and want to take another, you will need to put the previously held item down.
The name of the item you take must exactly match the name given in the task description. To find the right item, you can look in any possible places. You can start your search where the item is most likely to be found.

Please decide whether the previous action '{}' needs to be corrected and decide the most appropriate next action. If the previous action needs to be corrected, make sure the corrected action is different from the previous action and it is executable. The next action should be different from the corrected action. If the previous action does not need to be corrected, repeat the previous action as the corrected action.
Please reply in this format: "Previous action '{}' is [correct or wrong]. To fix this mistake, I should execute: [action].
The next action is: [action]."

Interact with a household to solve a task. Here are two examples.
<examples>
Here is the task
<new task description>
<current observation-action pairs>
\end{lstlisting}

\end{document}